\documentclass{article}
\usepackage{spconf,amsmath,graphicx}
\usepackage{multirow}
\usepackage{multicol}
\usepackage{enumitem}
\usepackage{cite} 
\usepackage{xcolor, colortbl}
\usepackage{tablefootnote}

\title{Exploiting prompt learning with pre-trained language models for Alzheimer's Disease detection}
\name{Yi Wang$^1$, Jiajun Deng$^1$, Tianzi Wang$^1$, Bo Zheng$^1$, Shoukang Hu$^2$, Xunying Liu$^1$, Helen Meng$^1$}
\address{$^1$The Chinese University of Hong Kong, Hong Kong SAR, China, \\
        $^2$Nanyang Technological University, Singapore}

\begin{document}
\ninept
\bstctlcite{IEEEexample:BSTcontrol}

\maketitle
\begin{abstract}
Early diagnosis of Alzheimer’s disease (AD) is crucial in facilitating preventive care and to delay further progression. Speech based automatic AD screening systems provide a non-intrusive and more scalable alternative to other clinical screening techniques. Textual embedding features produced by pre-trained language models (PLMs) such as BERT are widely used in such systems. However, PLM domain fine-tuning is commonly based on the masked word or sentence prediction costs that are inconsistent with the back-end AD detection task. To this end, this paper investigates the use of prompt-based fine-tuning of PLMs that consistently uses AD classification errors as the training objective function. Disfluency features based on hesitation or pause filler token frequencies are further incorporated into prompt phrases during PLM fine-tuning. 
{The decision voting based combination among systems using different PLMs (BERT and RoBERTa) or systems with different fine-tuning paradigms (conventional masked-language modelling fine-tuning and prompt-based fine-tuning) is further applied. }
Mean, standard deviation (std) and the maximum among accuracy scores over 15 experiment runs are adopted as performance measurements for the AD detection system. Mean detection accuracy of 84.20\% (with std 2.09\%, best 87.5\%) and 82.64\% (with std 4.0\%, best 89.58\%) were obtained using manual and ASR speech transcripts respectively on the ADReSS20 test set consisting of 48 elderly speakers\footnote{Code is available at https://github.com/yiwang454/prompt\_ad\_code}.
\end{abstract}
\vspace{-3mm}
\section{Introduction}
\vspace{-2mm}
\label{sec:intro}
Alzheimer’s disease (AD) is the most frequent form of dementia found in aged people. Its characteristics include progressive degradation of the memory, cognition, and motor skills, and consequently decline in the speech and language skills of patients\cite{sachdev2014classifying, vipperla2010ageing}. Currently, there is no effective cure for AD\cite{Yadav}, but an intervention approach applied in time can postpone its progression and reduce the negative impact on patients\cite{de2019brief}. Compared with other screening techniques, speech and language technology based automatic AD diagnosis is non-intrusive and more scalable. Therefore, there has been an increasing interest in the development of automatic AD screening and diagnosis systems, notably represented by the recent ADReSS challenges\cite{luz2020alzheimer, luz2021detecting}. 

Current speech-based AD detection systems use the following features: a) acoustic features extracted from the audio of AD assessment speech recordings; b) linguistic features extracted from the speech transcripts; or c) both a) and b). Commonly used forms of acoustic features include, but are not limited to, spectral features, prosodic features, vocal quality, and pre-trained deep neural network (DNN) embedding features\cite{balagopalan2020bert, syed2020automated, martinc2020tackling, syed2021tackling, rohanian2021alzheimer, pappagari2021automatic, pan2021using, jinchao2023leveraging}. Linguistic features include the handcrafted ones\cite{santos2017enriching, ammar2018speech, chien2018assessment, balagopalan2020bert, martinc2020tackling} capturing lexical or syntactical cues, or pre-trained text neural embeddings\cite{balagopalan2020bert, syed2020automated, baidu_paper, qiao2021alzheimer, jinchao2023leveraging}. Disfluency measures of speech were also found useful for AD detection tasks\cite{chien2018assessment, baidu_paper, qiao2021alzheimer, rohanian2021alzheimer}. To fully automate the AD detection process, ASR systems\cite{ammar2018speech, chien2018assessment, li2021comparative, qiao2021alzheimer, pan2021using, syed2021tackling, zhu2021wavbert, rohanian2021alzheimer, pappagari2021automatic, chen2019attention} have been increasingly often used to produce speech transcripts, in place of ground truth manual transcripts\cite{santos2017enriching, balagopalan2020bert, syed2020automated, baidu_paper, sarawgi2020multimodal, li2021comparative}. Previous work shows that among speech-based AD detection techniques, using linguistic features extracted from elderly subjects’ speech recording transcripts is particularly useful in differentiating between AD and non-AD labels, compared with using acoustic features alone \cite{balagopalan2020bert, syed2020automated, li2021comparative}. Among linguistic feature based methods, using pre-trained language models (PLM), either as feature extractors\cite{balagopalan2020bert, syed2020automated, li2021comparative, ye2021development, syed2021automated} or directly fine-tuning them on the AD/non-AD classification task\cite{baidu_paper, qiao2021alzheimer}, could achieve state-of-the-art detection accuracy {with manual transcripts and 93.8\% with ASR transcripts \cite{wang2022exploring} on the ADReSS20 dataset set. Meanwhile, \cite{martinc2021temporal} achieved a similarly competitive accuracy of 93.8\% with audio and text multimodal information using the temporal analysis method and 50-model majority voting}.

However, one critical challenge of applying PLM in AD detection is the inconsistency between the objective functions used during their pre-training and subsequent AD detection stage. The cost functions for pre-training transformer language models are usually based on cloze-style question answering tasks\cite{devlin2018bert, liu2019roberta, sun2019ernie} (e.g., word prediction via masked-language modelling), or the next sentence prediction task\cite{devlin2018bert}. Both learning objectives are different from that of the AD classification task. A commonly adopted practice in prior researches is based on a pipelined system architecture. Textual embedding features produced by PLMs that are offline fine-tuned via masked-language modelling (MLM) and sometimes together with sentence prediction tasks, before being fed into various forms of back-end classifiers optimized on AD detection, for example, the support vector machine (SVM) \cite{li2021comparative, ye2021development, wang2022exploring}. The learning cost function inconsistency between the front-end PLM feature extractor and the back-end AD classifiers remains unaddressed. Furthermore, such disjointed framework hinders the incorporation of additional fine-grained annotation labels that are useful for AD detection, for example, disfluency, age or educational background, into PLM training. 

To this end, we adopted a prompt-based fine-tuning paradigm to address the above inconsistency issues. By considering NLP problems as cloze-style tasks, prompt-based learning \cite{radford2019language, brown2020language, raffel2020exploring, schick2020exploiting}  has shown competitive performance in text classification tasks, especially in few-shot learning scenarios. So far, limited prior researches have studied its effectiveness on medical applications\cite{taylor2022clinical, sivarajkumar2022healthprompt}. In this paper, we first reformulate the AD versus non-AD classification problem into probabilistically predicting label words to fill in the prompt phrases, for example, based on the template “The diagnosis is $<$MASK$>$” , where “dementia” and “healthy” serve as possible labels for “$<$MASK$>$”. This allows the pre-trained BERT and RoBERTa language models to be consistently adapted in domain and fine-tuned with the AD classification task on the benchmark ADReSS2020 training data \cite{luz2020alzheimer} (a.k.a. cookie task of Dementia Pitt corpus \cite{becker1994natural}). To mitigate the risk of overfitting during prompt-based PLM fine-tuning on limited AD detection training data, model averaging over different update intervals using decision voting was used. Second, disfluency features based on hesitation or pause filler token frequencies are further incorporated into prompt phrases during PLM fine-tuning for AD detection. Finally, in order to explore the complementarity between BERT or RoBERTa based PLMs that are either fine-tuned with prompt learning, or optimized using conventional masked word or sentence prediction costs, decision voting based system combination between them is further applied. Experiments are conducted on the ADReSS20 Challenge dataset\cite{luz2020alzheimer}.

The main contributions of this paper are summarized below: 1) it presents the first work adopting prompt learning based PLM fine-tuning for  automatic AD detection. In contrast, prior researches on prompt learning for medical application were not conducted for AD detection but for clinical tasks with different characteristics \cite{taylor2022clinical, sivarajkumar2022healthprompt}. 2) the proposed prompt learning based approaches allow additional fine-grained AD detection relevant features such as disfluency to be further incorporated by simply adding a relevant description to the prompt phrase. In contrast, this is not feasible when the conventional pipelined methods are applied.
3) it achieves mean AD detection accuracies of 84.20\% (with std 2.09\%, best 87.5\%) obtained using manual transcripts and 82.64\% (with std 4.0\%, best 89.58\%) using ASR transcripts, with the best accuracies among both systems reaching state-of-the-art level. It also shows a notably minor variance in model performance than our previous work\cite{wang2022exploring}, which obtained 81.09\% (with std 3.95\%, best 91.67\%) and 81.74\% (with std 4.47\%, best 93.75\%) using manual transcripts and ASR transcripts.

\vspace{-3mm}
\section{Task Description}
\label{sec:TaskDescription}
\vspace{-3mm}
This section describes the text data used in this paper and the baseline system structure.
\vspace{-4mm}
\subsection{Text Data}
\vspace{-0.2cm}
In this paper, AD detection system training and evaluation are conducted on the ADReSS20 Challenge dataset \cite{luz2020alzheimer}. It is selected from the Cookie Theft picture description part of the Pitt Corpus in the DementiaBank database \cite{becker1994natural} with balanced age, gender and AD label distribution among participants. The data for each participant includes one speech recording of Cookie Theft storytelling, corresponding manual transcripts annotated by the CHAT coding system \cite{macwhinney2000childes} and a binary AD label. Both the participant and the interviewer parts are provided, but only the participant part is used in our experiments. For each participant, all speech segments are combined as the input text. The dataset is divided into the training set from 108 subjects and the test set from 48 subjects, corresponding to 2 hours 9 minutes and 1 hour 6 minutes of recordings respectively.

We conduct our experiments with either ground truth manual transcripts or transcripts generated by ASR system from audios.
\\
\textbf{ASR systems:} The experimental results in \cite{wang2022exploring} suggest that the transcripts generated by the adapted hybrid CNN-TDNN ASR system \cite{ye2021development} achieve better AD detection performance than those obtained from the adapted E2E Conformer model \cite{tianzi2022conformer}. Hence, the hybrid CNN-TDNN ASR system is used.

\vspace{-4mm}
\subsection{Baseline System}
\label{sec:baseline}
\vspace{-2mm}
We take an existing pipelined architecture \cite{wang2022exploring} as our baseline AD detection system. As is shown in Figure 1(A right), it consists of PLM text embedding extractors and back-end classifiers producing detection decisions from text embeddings, accompanied by majority voting among subsystems with different PLM or classifier components. The PLM text encoders are fine-tuned with MLM (and Next Sentence Prediction for BERT) on the ADReSS training data, as shown in Figure 1(A left). Since the fine-tuning objective is MLM instead of AD classification, the detection accuracy scores at consecutive epochs fluctuate. Models at the final three update epochs during the 30-epoch fine-tuning were used to produce separate text embedding features for back-end classifiers. Their AD detection outputs are combined by majority voting to reduce the risk of over-fitting and smooth the unstable performance. SVM is used as the back-end classifier, as it outperformed the other classifiers. The systems using different PLMs as the feature extractor show complementarity. Therefore, model combination by majority voting among the systems is applied to improve performances, reaching state-of-the-art ADReSS test accuracy of 87.5\%. 
\\
\textbf{PLMs:} BERT or RoBERTa models are taken as the PLMs for the pipelined PLM embedding extractor and classifier architecture and further for the prompt-based fine-tuning method, as their effectiveness in the AD detection task has been proven in \cite{wang2022exploring, syed2021automated}. The BERT pre-trained model ${\tt bert-base-uncased}$  is based on the open-source model library\footnote{https://huggingface.co/bert-base-uncased}, with the standard BERT tokenizer. This paper also explored the use of ${\tt roberta-base}$ model and the accompanying tokenizer\footnote{https://huggingface.co/roberta-base}. The max input length of both PLMs is 512.
\\We evaluated our models with 10-fold cross validation (CV) on the training set and tested on the test set. The evaluation of each system is conducted 15 times by using 15 different random seeds for Pytorch random initialization setting\footnote{https://pytorch.org/docs/stable/notes/randomness.html}. Mean, standard deviation (std) and the maximum among accuracy scores of all runs are used as performance measures. 

\begin{figure*}[t]
  \centering
  \includegraphics[width=\linewidth]{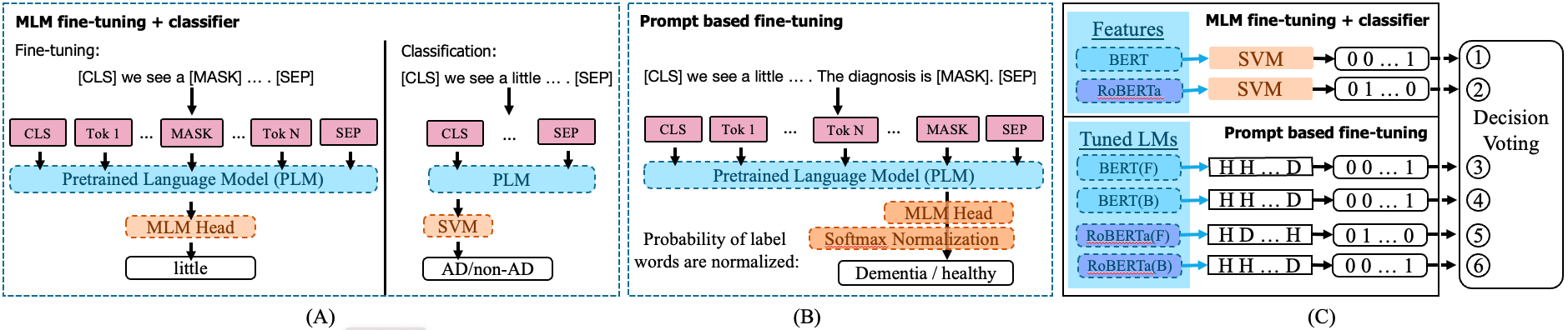}
  \vspace{-5mm}
  \caption{Example of (A) the baseline AD detection system consisting of MLM fine-tuned text embedding extractors and back-end classifier (denoted by MLM fine-tuning + classifier), (B) the prompt-based fine-tuned PLM detection system (C) possible model combination ways among systems of MLM fine-tuned / prompt-based fine-tuned BERT/RoBERTa, with prompt phrase positioned at the front (denoted by “F”) / back (denoted by “B”) of input texts, conducted by majority voting among AD decisions.}
  \vspace{-0.2cm}
  \label{fig:AD_detection}
\vspace{-0.2cm}
\end{figure*}

\vspace{-3mm}
\section{Prompt-based Fine-tuning}
\label{sec:PromptTuning}
\vspace{-3mm}
We conduct prompt-based PLM fine-tuning using manually designed prompt phrases. We reformulate the AD versus non-AD classification problem into probabilistically predicting the label word to fill in the masked position, for example, based on the general prompt template “The diagnosis is $<$MASK$>$.”, where “dementia” and “healthy” serve as the AD and non-AD labels for “$<$MASK$>$” respectively. Each label word corresponds to one token in the vocabulary of BERT or RoBERTa based PLMs. The speech transcripts are concatenated with the prompt phrase for PLMs to predict logits representing the probablities for the AD and non-AD label words to fill in the masked position given the corresponding vocabulary. The resulting logits are then normalized via a Softmax layer, to compute binary cross entropy loss on AD detection over the logits for the label words “dementia” and “healthy”
as shown in Figure 1(B). Separate back-end classifiers included in the pipelined detection systems are no longer required. Different positions to insert the prompt phrases, either at the front or back of the input text, are also explored. This serves to measure the effect of different prompt locations on the resulting attention based PLM contextual representations of labelled input text.

We implemented the prompt-based PLM fine-tuning paradigm training and evaluation using the OpenPrompt framework\cite{ding2021openprompt} based on Pytorch. The hyper-parameters for prompt-based fine-tuning are optimized by greedy search on CV using the mean accuracy score as the scheduling criterion. Prompt-based fine-tuning of 10 epochs was performed on all the parameters of both BERT and RoBERTa PLMs. The learning rate setting of 1e-05, batch size of 1 and the AdamW optimizer were used. A weight decay 0.01 was applied to the LayerNorm modules. In common with the non-prompt, MLM based fine-tuning, AD detection outputs obtained at the last three epochs during fine-tuning were majority voted upon to generate the final decision.
\begin{table}[t]
\setlength{\abovecaptionskip}{0cm}
\centering
\scalebox{0.67}{
\begin{tabular}{c|c|c|c|ccc|ccc}
\noalign{\hrule height 1.0pt}
\multirow{2}{*}{Sys.} &
  \multirow{2}{*}{\begin{tabular}[c]{@{}c@{}}PLMs\end{tabular}} &
  \multirow{2}{*}{\begin{tabular}[c]{@{}c@{}}Fine-tuning\\ Paradigm\end{tabular}} &
  \multirow{2}{*}{\begin{tabular}[c]{@{}c@{}}Prompt\\ location\end{tabular}} &
  \multicolumn{3}{c|}{CV Acc (\%)} &
  \multicolumn{3}{c}{Test Acc (\%)} \\ \cline{5-10} 
 &
   &
   &
   &
  \multicolumn{1}{c|}{Mean} &
  \multicolumn{1}{c|}{Std} &
  Best &
  \multicolumn{1}{c|}{Mean} &
  \multicolumn{1}{c|}{Std} &
  Best \\ \hline
1 &
  \multirow{3}{*}{BERT} &
  MLM,NSP &
  N/A &
  \multicolumn{1}{c|}{81.30} &
  \multicolumn{1}{c|}{2.58} &
  85.19 &
  \multicolumn{1}{c|}{78.05} &
  \multicolumn{1}{c|}{3.79} &
  85.42 \\  \cline{3-10} 
2 &
   &
  \multirow{2}{*}{Prompt} &
  Front &
  \multicolumn{1}{c|}{81.30} &
  \multicolumn{1}{c|}{1.03} &
  83.33 &
  \multicolumn{1}{c|}{79.58} &
  \multicolumn{1}{c|}{2.87} &
  85.42 \\  \cline{4-10} 
3 &
   &
   &
  Back &
  \multicolumn{1}{c|}{80.43} &
  \multicolumn{1}{c|}{1.72} &
  83.33 &
  \multicolumn{1}{c|}{81.25} &
  \multicolumn{1}{c|}{2.28} &
  83.33  \\ \hline
4 &
  \multirow{3}{*}{RoBERTa} &
  MLM &
  N/A &
  \multicolumn{1}{c|}{76.17} &
  \multicolumn{1}{c|}{2.96} &
  83.33 &
  \multicolumn{1}{c|}{80.00} &
  \multicolumn{1}{c|}{4.08} &
  85.42 \\  \cline{3-10} 
5 &
   &
  \multirow{2}{*}{Prompt} &
  Front &
  \multicolumn{1}{c|}{82.16} &
  \multicolumn{1}{c|}{2.54} &
  86.11 &
  \multicolumn{1}{c|}{82.08} &
  \multicolumn{1}{c|}{2.72} &
  85.42  \\  \cline{4-10} 
6 &
   &
   &
  Back &
 \multicolumn{1}{c|}{83.40} &
  \multicolumn{1}{c|}{2.50} &
  87.04 &
  \multicolumn{1}{c|}{80.97} &
  \multicolumn{1}{c|}{3.48} &
  85.42 \\ \noalign{\hrule height 1.0pt}
\end{tabular}}
\vspace{-2mm}
\caption{AD detection performance, in terms of the mean, standard deviation (std) and best of acuuracy scores, evaluated by 10-fold CV on the ADReSS2020\cite{luz2020alzheimer} training set and test set with ground truth transcript using either MLM or prompt based fine-tuning, BERT and RoBERTa are used as PLMs. Sys.2\&5 and Sys.3\&6 show results of the prompt-based fine-tuned PLMs that position the prompt phrases at the back and the front of the input text respectively.}
\vspace{-5.5mm}
\end{table}

The performance of AD detection systems using prompt-based fine-tuned PLMs evaluated on the training data with 10-fold CV and the test set are shown in Table 1. Several main trends can be observed. Firstly, the prompt-based fine-tuned PLMs show similar\footnote{Accuracy difference less than 1\% on CV, 2\% on the test set is regarded as similar, corresponding to one participant}
(Sys.2-3 vs. Sys.1 on CV) or better performance (Sys.2-3 vs. Sys.1 on Test, Sys.5-6 vs. Sys.4 on CV and Test) compared to those of the baseline MLM fine-tuning PLMs in terms of the mean accuracy. Secondly, the prompt-based fine-tuned PLMs produce more consistent results than the MLM fine-tuning PLMs, reflected by smaller standard deviation (std) scores of the prompt-based fine-tuned PLMs than the baselines (Sys.2-3 vs. Sys.1, Sys.5-6 vs. Sys.4). Thirdly, no significant accuracy difference is obtained between the prompt trained PLMs that position the prompt phrases at the back and the front of the input text (Sys.2 vs. Sys.3, Sys.5 vs. Sys.6).

\vspace{-3mm}
\section{Model Combination }
\label{sec:ModelCombination}
\vspace{-2mm}
\subsection{Combining MLM and Prompt-based PLM Fine-tuning}
\label{sec:prompt_cls}
\vspace{-2mm}
As mentioned above, different prompt phrase locations may guide the systems to capture different information. In addition, different fine-tuning paradigms can also allow PLMs to learn linguistic information via diverse embeddings. To exploit such diversity among different systems, late fusion by majority voting is conducted over decisions made by systems with the front and the back positioned prompt phrase (i.e., the fusion of (3) \& (4) or (5) \& (6) shown in Fig. 1(C)). A similar and wider scope of decision voting is further explored among the front prompt and back prompt fine-tuned BERT/RoBERTa PLMs, together with the MLM fine-tuned system (i.e., the fusion of (1), (3) and (4) or (2), (5) and (6) in Fig. 1(C)). 

As shown in Table 2 (column 5-10 for CV and Test set using manual transcripts), the first general trend is that the combination of front and back prompt-based fine-tuned PLMs shows no further performance improvements over any individual models in terms of the mean, std and best accuracy scores across CV and test sets (Sys.4 vs. Sys.2-3, Sys.9 vs. Sys.7-8). These results suggest altering the position of inserting the prompt phrases produced insufficient diversity among the resulting PLMs. Secondly, the three-way combination of MLM and two prompt-based fine-tuned PLMs consistently outperform each individual models in terms of the mean scores by 7.84\% / 4.03\% (CV/test) at most, and the std scores by 1.35\% / 1.96\% (CV/test) at most (Sys.5 vs. Sys.1-4, Sys.10 vs. Sys.6-9). Despite the slight degradation in the best accuracy for BERT on the test set, the combination of systems with different fine-tuning paradigms can improve the AD detection accuracy in terms of both mean and std. 

\begin{table*}[htbp]
\centering
\scalebox{0.7}{
\centering
\begin{tabular}{c|c|c|c|ccc|ccc|ccc|ccc|ccc}
\noalign{\hrule height 1.0pt}
\multirow{2}{*}{Sys.} &
  \multirow{2}{*}{PLMs} &
  \multirow{2}{*}{Fine-tuning Paradigm} &
  \multirow{2}{*}{\begin{tabular}[c]{@{}c@{}}Combi\\-nation\end{tabular}} &
    \multicolumn{3}{c|}{\begin{tabular}[c]{@{}c@{}}CV Acc (\%)\\ Manual\end{tabular}} &
  \multicolumn{3}{c|}{\begin{tabular}[c]{@{}c@{}}Test Acc (\%)\\ Manual\end{tabular}} &
  \multicolumn{3}{c|}{\begin{tabular}[c]{@{}c@{}}Test  Acc  (\%)\\   Manual+disfl.\end{tabular}} &
  \multicolumn{3}{c|}{\begin{tabular}[c]{@{}c@{}}Test Acc (\%) \\ ASR\end{tabular}} &
  \multicolumn{3}{c}{\begin{tabular}[c]{@{}c@{}}Test  Acc  (\%)\\ ASR+disfl.\end{tabular}} \\ \cline{5-19}
 &
   &
   &
   &
  \multicolumn{1}{c|}{Mean} &
  \multicolumn{1}{c|}{Std} &
  Best &
  \multicolumn{1}{c|}{Mean} &
  \multicolumn{1}{c|}{Std} &
  Best &
  \multicolumn{1}{c|}{Mean} &
  \multicolumn{1}{c|}{Std} &
  Best &
  \multicolumn{1}{c|}{Mean} &
  \multicolumn{1}{c|}{Std} &
  Best &
  \multicolumn{1}{c|}{Mean} &
  \multicolumn{1}{c|}{Std} &
  Best \\ \noalign{\hrule height 1.0pt}
1 &
  \multirow{5}{*}{BERT} &
  MLM & \multirow{3}{*}{$\times$}
   & 
  \multicolumn{1}{c|}{81.30} &
  \multicolumn{1}{c|}{2.58} &
  85.19 &
  \multicolumn{1}{c|}{78.05} &
  \multicolumn{1}{c|}{3.79} &
  85.42 &
  \multicolumn{3}{c|}{-} &
  \multicolumn{1}{c|}{78.61} &
  \multicolumn{1}{c|}{5.02} &
  87.50 &
  \multicolumn{3}{c}{-} \\   \cline{3-3} \cline{5-19} 
2 &
   &
  Prompt (Front) &
   &
  \multicolumn{1}{c|}{81.30} &
  \multicolumn{1}{c|}{1.03} &
  83.33 &
  \multicolumn{1}{c|}{79.58} &
  \multicolumn{1}{c|}{2.87} &
  85.42 &
  \multicolumn{1}{c|}{80.28} &
  \multicolumn{1}{c|}{3.22} &
  85.42 &
  \multicolumn{1}{c|}{77.08} &
  \multicolumn{1}{c|}{4.75} &
  85.42 &
  \multicolumn{1}{c|}{\bf{80.69}} &
  \multicolumn{1}{c|}{3.44} &
  85.42 \\  \cline{3-3} \cline{5-19} 
3 &
   &
  Prompt (Back) &
   &
  \multicolumn{1}{c|}{80.43} &
  \multicolumn{1}{c|}{1.72} &
  83.33 &
  \multicolumn{1}{c|}{81.25} &
  \multicolumn{1}{c|}{2.28} &
  83.33 &
   \multicolumn{1}{c|}{81.94} &
  \multicolumn{1}{c|}{2.48} &
  85.42 &
  \multicolumn{1}{c|}{80.14} &
  \multicolumn{1}{c|}{3.94} &
  87.50 &
  \multicolumn{1}{c|}{77.50} &
  \multicolumn{1}{c|}{4.38} &
  85.42  \\  \cline{3-19} 
4 &
   &
  Prompt (Front + Back) & \multirow{2}{*}{$\surd$}
   &
  \multicolumn{1}{c|}{80.74} &
  \multicolumn{1}{c|}{1.32} &
  83.33 &
  \multicolumn{1}{c|}{82.22} &
  \multicolumn{1}{c|}{1.99} &
  85.42 &
  \multicolumn{1}{c|}{82.50} &
  \multicolumn{1}{c|}{1.98} &
  85.42 &
  \multicolumn{1}{c|}{74.72} &
  \multicolumn{1}{c|}{5.15} &
  83.33 &
  \multicolumn{1}{c|}{75.56} &
  \multicolumn{1}{c|}{4.78} &
  85.42  \\  \cline{3-3} \cline{5-19} 
5 &
   &
  MLM + Prompt (Front + Back) &
   &
  \multicolumn{1}{c|}{83.52} &
  \multicolumn{1}{c|}{1.23} &
  85.19 &
  \multicolumn{1}{c|}{82.08} &
  \multicolumn{1}{c|}{1.83} &
  83.33 &
  \multicolumn{3}{c|}{-} &
  \multicolumn{1}{c|}{81.11} &
  \multicolumn{1}{c|}{3.69} &
  87.50 &
  \multicolumn{3}{c}{-} \\ \noalign{\hrule height 1.0pt}
6 &
  \multirow{5}{*}{RoBERTa} &
  MLM & \multirow{3}{*}{$\times$}
   &
  \multicolumn{1}{c|}{76.17} &
  \multicolumn{1}{c|}{2.96} &
  83.33 &
  \multicolumn{1}{c|}{80.00} &
  \multicolumn{1}{c|}{4.08} &
  85.42 &
  \multicolumn{3}{c|}{-} &
  \multicolumn{1}{c|}{77.64} &
  \multicolumn{1}{c|}{5.30} &
  87.50 &
  \multicolumn{3}{c}{-} \\ \cline{3-3} \cline{5-19} 
7 &
   &
  Prompt (Front) &
   &
  \multicolumn{1}{c|}{82.16} &
  \multicolumn{1}{c|}{2.54} &
  86.11 &
  \multicolumn{1}{c|}{82.08} &
  \multicolumn{1}{c|}{2.72} &
  85.42 &
  \multicolumn{1}{c|}{78.89} &
  \multicolumn{1}{c|}{3.79} &
  85.42 &
  \multicolumn{1}{c|}{79.44} &
  \multicolumn{1}{c|}{4.29} &
  87.50 &
  \multicolumn{1}{c|}{79.86} &
  \multicolumn{1}{c|}{3.20} &
  85.42 \\  \cline{3-3} \cline{5-19} 
8 &
   &
  Prompt (Back) &
   &
  \multicolumn{1}{c|}{83.40} &
  \multicolumn{1}{c|}{2.50} &
  87.04 &
  \multicolumn{1}{c|}{80.97} &
  \multicolumn{1}{c|}{3.48} &
  85.42 &
  \multicolumn{1}{c|}{81.25} &
  \multicolumn{1}{c|}{3.04} &
  85.42 &
  \multicolumn{1}{c|}{78.89} &
  \multicolumn{1}{c|}{4.80} &
  85.42 &
  \multicolumn{1}{c|}{76.67} &
  \multicolumn{1}{c|}{4.04} &
  85.42 \\  \cline{3-19} 
9 &
   &
  Prompt (Front + Back) & \multirow{2}{*}{$\surd$}
   &
  \multicolumn{1}{c|}{83.02} &
  \multicolumn{1}{c|}{2.08} &
  86.11 &
  \multicolumn{1}{c|}{82.08} &
  \multicolumn{1}{c|}{2.61} &
  87.50 &
  \multicolumn{1}{c|}{81.81} &
  \multicolumn{1}{c|}{2.21} &
  85.42 &
  \multicolumn{1}{c|}{75.83} &
  \multicolumn{1}{c|}{3.94} &
  81.25 &
  \multicolumn{1}{c|}{75.69} &
  \multicolumn{1}{c|}{3.11} &
  81.25 \\  \cline{3-3} \cline{5-19} 
10 &
   &
  MLM + Prompt (Front + Back) &
   &
  \multicolumn{1}{c|}{84.01} &
  \multicolumn{1}{c|}{2.04} &
  87.96 &
  \multicolumn{1}{c|}{83.61} &
  \multicolumn{1}{c|}{2.13} &
  87.50 &
  \multicolumn{3}{c|}{-} &
  \multicolumn{1}{c|}{\bf{82.64}} &
  \multicolumn{1}{c|}{4.00} &
  89.58 &
  \multicolumn{3}{c}{-} \\\noalign{\hrule height 1.0pt}
11 &
  \multirow{5}{*}{\begin{tabular}[c]{@{}c@{}}Bert+\\ RoBERTa\end{tabular}} &
  MLM & \multirow{5}{*}{$\surd$}
   &
  \multicolumn{1}{c|}{81.39} &
  \multicolumn{1}{c|}{2.84} &
  88.89 &
  \multicolumn{1}{c|}{81.09} &
  \multicolumn{1}{c|}{3.95} &
  91.67 &
  \multicolumn{3}{c|}{-} &
  \multicolumn{1}{c|}{81.74} &
  \multicolumn{1}{c|}{4.47} &
  93.75 &
  \multicolumn{3}{c}{-} \\ \cline{3-3} \cline{5-19} 
12 &
   &
  Prompt (Front) &
   &
  \multicolumn{1}{c|}{81.71} &
  \multicolumn{1}{c|}{1.52} &
  85.19 &
  \multicolumn{1}{c|}{81.69} &
  \multicolumn{1}{c|}{2.53} &
  87.50 &
   \multicolumn{1}{c|}{82.20} &
  \multicolumn{1}{c|}{2.17} &
  87.50 &
  \multicolumn{1}{c|}{74.67} &
  \multicolumn{1}{c|}{4.63} &
  85.42 &
  \multicolumn{1}{c|}{78.19} &
  \multicolumn{1}{c|}{3.94} &
  85.42 \\  \cline{3-3} \cline{5-19} 
13 &
   &
  Prompt (Back) &
   &
  \multicolumn{1}{c|}{81.40} &
  \multicolumn{1}{c|}{1.68} &
  84.26 &
  \multicolumn{1}{c|}{81.22} &
  \multicolumn{1}{c|}{2.39} &
  85.42 &
  \multicolumn{1}{c|}{82.36} &
  \multicolumn{1}{c|}{2.48} &
  87.50 &
  \multicolumn{1}{c|}{75.65} &
  \multicolumn{1}{c|}{3.56} &
  87.50 &
  \multicolumn{1}{c|}{73.33} &
  \multicolumn{1}{c|}{3.97} &
  79.17 \\  \cline{3-3} \cline{5-19} 
14 &
   &
  Prompt (Front + Back) & 
   &
  \multicolumn{1}{c|}{83.32} &
  \multicolumn{1}{c|}{1.34} &
  87.04 &
  \multicolumn{1}{c|}{83.10} &
  \multicolumn{1}{c|}{2.17} &
  87.50 &
  \multicolumn{1}{c|}{\bf{83.32}} &
  \multicolumn{1}{c|}{2.15} &
  87.50 &
  \multicolumn{1}{c|}{78.01} &
  \multicolumn{1}{c|}{3.45} &
  85.42 &
  \multicolumn{1}{c|}{79.31} &
  \multicolumn{1}{c|}{3.52} &
  83.33 \\ \cline{3-3} \cline{5-19} 
15 &
   &
  MLM + Prompt (Front + Back) &
   &
  \multicolumn{1}{c|}{\bf{84.20}} &
  \multicolumn{1}{c|}{1.49} &
  87.04 &
  \multicolumn{1}{c|}{\bf{84.20}} &
  \multicolumn{1}{c|}{2.09} &
  87.50 &
  \multicolumn{3}{c|}{-} &
  \multicolumn{1}{c|}{81.56} &
  \multicolumn{1}{c|}{3.09} &
  87.50 &
  \multicolumn{3}{c}{-} \\ \noalign{\hrule height 1.0pt}
\end{tabular}}
\vspace{-2mm}
\caption{AD detection performance of the baseline MLM fine-tuned PLMs\cite{wang2022exploring}, the front and the back positioned prompt-based fine-tuned PLMs, the combination of system with different fine-tuning paradigms or different base PLMs are shown. All systems evaluated with manual transcripts (Manual) on CV; Manual, ASR transcripts (ASR), Manual with disfluency information (“+disfl.”) or ASR+disfl. on the test set.}
\vspace{-5.5mm}
\label{tab:prompt_baseline}
\end{table*}
\vspace{-3mm}
\subsection{Combining BERT and RoBERTa}
\label{sec:bert_roberta}
\vspace{-2mm}
Previous research \cite{wang2022exploring} demonstrated that decision voting from AD detection systems using BERT and RoBERTa PLMs extracted text embedding features can be further combined to exploit their complementarity. To this end, we conducted late fusion by majority voting over the decisions made by BERT and Roberta PLMs fine-tuned using either MLM (Fig. 1(C) (1)+(2)), front and/or back positioned prompts alone (Fig. 1(C) (3)+(4), (5)+(6)), or (3)+(4)+(5)+(6)), or larger ensemble involving both (as Figure 1C (3)$+$(4)$+$(5)$+$(6), (1)$+$(2)$+$(3)$+$(4)$+$(5)$+$(6)). Permutation of 15 random initialization seeds for any of the two single PLMs being combined lead to a total of $15^2=255$ decisions, among which the mean, std and best AD detection accuracy scores are calculated to measure the performance of combined systems.

Several trends can be found in the BERT plus RoBERTa PLM combination results in Table 2 (col. 5-10; Sys.11-15 for CV and Test set using manual transcript). First, improved mean accuracy scores are obtained using a range of BERT plus RoBERTa PLM combination experiments including the baseline MLM fine-tuned PLMs (Sys.11 vs. Sys.1\&6), the combination system between the front plus back prompt fine-tuned PLMs (Sys.14 vs. Sys.4\&9), and those using either MLM or prompt-based fine-tuning (Sys.15 vs. Sys.5\&10). Secondly, no consistent BERT plus RoBERTa model combination improvement was obtained among the PLMs fine-tuned using front or back positioned prompts alone (Sys.12 vs. Sys.2\&7, Sys.13 vs. Sys.3\&8). This indicates there is insufficient cross system diversity among these to be exploited in combination. Thirdly, combining the baseline MLM fine-tuned BERT and RoBERTa models yields a noticeable increase in the best accuracy score over the single PLM system. This best accuracy yet present a larger deviation from the corresponding mean score (Sys.11). In contrast, the best preformed combination systems involving prompt-based fine-tuned PLMs into the combination (Sys.15) generally have smaller std scores over the MLM fine-tuned BERT and RoBERTa combination models, indicating increased consistency in performance across different random initialization seeds.

\vspace{-3mm}
\section{Disfluency Features}
\label{sec:Disfluency}
\vspace{-3mm}
To exploit disfluency features based on interjection, hesitation, or pause filler tokens in the AD detection task, we incorporate transcription-based disfluency detection task into prompt-based fine-tuning by concatenating the disfluency prompt phrase \textbf{“Speech is $<$MASK$>$.”} in front of the “Diagnosis” phrase, resulting in the multi-task prompt phrase \textbf{“Speech is $<$MASK$>$. Diagnosis is $<$MASK$>$.”}. 
Subjects whose disfluency token frequency is no less than the threshold are assigned to the label “stumbling”, and otherwise “fluent”. Multi-task fine-tuning is conducted using the interpolated AD+disfluency ({1:1} weighting) classification cost. 

The disfluency tokens in the manual transcripts include interjection words like “uh”, “hm”; pause symbols like  (.), (..); or subject actions like “laughs”, “clear throat” which also indicate hesitations. We empirically set the threshold $=11$ for manual transcript based disfluency features by selecting the value that maximizes the correlation between two binary categorizations AD/non-AD and Stumbling/Fluent on the training set. Threshold based on manual transcripts splits the training set into 22 / 86 participants and the test set into 13 / 35 participants for Stumbling/Fluent categories. However, among the above three kinds of tokens, only interjection words can be transcribed by ASR systems, leading to different distributions of disfluency frequencies on the manual and the ASR transcripts. To implement disfluency detection task with ASR transcripts, we 
use an empirically set threshold $=4$ that produces the split closest to the manual based training set split, resulting in the Stumbling/Fluent split of 9 / 39.

The performance of prompt-based fine-tuned PLMs and their combination after incorporating disfluency features are shown in Table 2 ({col.} 11-13; Sys.2-4, 7-9, 12-14). Adding the disfluency task did not produce consistent performance improvements over the AD detection single-task fine-tuned systems in terms of mean, std and best scores. This is probably for the reason that hesitation symbols in speech transcripts are already captured by PLMs in AD detection single-task fine-tuned systems. Incorporating additional disfluency features that are outside the speech utterance boundaries, and not considered by the ASR systems, will be studied in future research. 

\vspace{-3mm}
\section{ASR Output Based AD Detection}
\label{sec:ASR_AD}
\vspace{-3mm}
In this Section, the performance of various baseline MLM, or prompt-based fine-tuned PLMs with or without disfluency features, and their combination described in Sections 2-5 are evaluated using ASR system transcripts.
Several trends can be found in Table 2 (col.14-19). First, regarding the mean accuracy, using ASR transcripts leads to performance degradation compared with using manual transcripts (col.14 vs. col.8), as observed in the experiment results of the prompt-based fine-tuned PLMs (Sys.2-5, Sys.7-10) and those combining MLM and prompt fine-tuned PLMs (Sys.12-15). In contrast, the MLM fine-tuned systems generally produce comparable mean accuracy scores using manual or ASR transcripts (Sys.1\&11), with the exception of the mean accuracy of RoBERTa based system (Sys.6).
Secondly, larger accuracy std scores are found among most systems using the errorful ASR than manual transcripts (col.15 vs.col.9). 
Thirdly, the combination over prompt-based fine-tuned PLMs with different prompt locations failed to produce consistent performance gains (col.14-16; Sys.4 vs.Sys.2-3, Sys.9 vs.Sys.7-8, Sys.14 vs. Sys.12-13). 
Finally, similar trends as discussed above are observed after integrating disfluency features in prompt based PLM fine-tuning (col.17-19; Sys.2-4, 7-9, 12-14). 

\vspace{-3mm}
\section{Conclusion}
\label{sec:Conclusion}
\vspace{-3mm}
Prompt learning based pre-trained language model (PLM) fine-tuning approaches for automatic AD detection is investigated in this paper to consistently use minimizing AD classification errors as the training objective function. Disfluency features based on pause filler token frequencies are further incorporated into prompt phrases during PLM fine-tuning. 
The complementarity between BERT or RoBERTa PLMs that are either fine-tuned with prompt learning, or optimized using conventional masked word or sentence prediction costs, is further exploited using system combination.
Experiments conducted on the ADReSS20 Challenge dataset suggest that improved AD detection mean accuracy and reduced standard deviation scores
are produced using prompt-based fine-tuned PLMs and their combination. State-of-the-art AD detection results of mean accuracy 84.20\% (with std 2.09\%, best 87.5\%) and 82.64\% (with std 4.0\%, best 89.58\%) were obtained using manual and ASR speech transcripts respectively on the ADReSS20 test set. Future research will focus on reducing the performance fragility to ASR transcript errors and to integrate richer disfluecny features from speech audio. 
\vspace{-0.3cm}
\section{Acknowledgements}
\vspace{-0.3cm}
\label{ssec:a}
This research is supported by Hong Kong RGC GRF grant No. 14200021, 14200218, 14200220, TRS T45-407/19N and Innovation \& Technology Fund grant No. ITS/254/19, ITS/218/21.



\bibliographystyle{IEEEtran}
\bibliography{strings,refs}

\end{document}